\pdfoutput=1

\documentclass[11pt]{article}

\usepackage[]{acl}

\usepackage{times}
\usepackage{latexsym}

\usepackage{algorithm}
\usepackage{multirow}
\usepackage{graphicx}
\usepackage{subfigure}
\usepackage{amsmath}
\usepackage{enumitem}
\usepackage{booktabs}
\usepackage{xspace}
\usepackage{microtype}
\usepackage{multicol}
\usepackage{caption}
\usepackage{amssymb}
\usepackage{array}
\usepackage{fp}
\usepackage{makecell}
\usepackage{flushend}
\usepackage{cases}
\usepackage{color}
\usepackage{algpseudocode}
\usepackage{listings}
\usepackage{xcolor}
\usepackage{mathrsfs}

\newcommand{\paratitle}[1]{\vspace{1.5ex}\noindent\textbf{#1}}
\newcommand{\ie}{\emph{i.e.,}\xspace}

\newcommand{\eg}{\emph{e.g.,}\xspace}

\newcommand{\ignore}[1]{}

\usepackage[T1]{fontenc}

\usepackage[utf8]{inputenc}

\usepackage{microtype}

\title{\emph{Great Truths are Always Simple}: A Rather Simple Knowledge Encoder for Enhancing the Commonsense Reasoning Capacity of Pre-Trained Models}

\author{
    \textbf{Jinhao Jiang\textsuperscript{{1},{3}}\thanks{\llap{}\:\:\:Equal contributions. },
	        Kun Zhou\textsuperscript{{2},{3}}\footnotemark[1],
	        Wayne Xin Zhao\textsuperscript{{1},{3},{4}}\thanks{\llap{}\:\:\:Corresponding authors. } ~\and
	        Ji-Rong Wen\textsuperscript{{1},{2},{3}}}\\
	\textsuperscript{1}Gaoling School of Artificial Intelligence, Renmin University of China.\\
	\textsuperscript{2}School of Information, Renmin University of China.\\
	\textsuperscript{3}Beijing Key Laboratory of Big Data Management and Analysis Methods.\\
	\textsuperscript{4}Beijing Academy of Artificial Intelligence, Beijing, 100084, China.\\
    	\texttt{jiangjinhao@ruc.edu.cn, francis\_kun\_zhou@163.com,} \\
	\texttt{batmanfly@gmail.com, jrwen@ruc.edu.cn}
}

\begin{document}
\maketitle
\begin{abstract}
Commonsense reasoning in natural language is a desired ability of artificial intelligent systems. 
For solving complex commonsense reasoning tasks, a typical solution is to enhance pre-trained language models~(PTMs) with a knowledge-aware graph neural network~(GNN) encoder that models a commonsense knowledge graph~(CSKG).
Despite the effectiveness, these approaches are built on heavy architectures, and can't clearly explain how external knowledge resources improve the reasoning capacity of PTMs. 
Considering this issue, we conduct a deep empirical analysis, and find that it is indeed \emph{relation features} from CSKGs (but not \emph{node features}) that mainly contribute to the performance improvement of PTMs. 
Based on this finding, we design a simple MLP-based knowledge encoder that utilizes statistical relation paths as features. 
Extensive experiments conducted on five benchmarks demonstrate the effectiveness of our approach, which also largely reduces the parameters for encoding CSKGs.
Our codes and data are publicly available at~\url{https://github.com/RUCAIBox/SAFE}.
\end{abstract}

\section{Introduction} \label{introduction}
In the era of artificial intelligence, it is desirable that intelligent systems can be empowered by the capacity of commonsense reasoning in natural language. 
For this purpose, a surge of commonsense reasoning tasks and datasets are proposed to evaluate and improve such an ability of NLP models, \eg CommonsenseQA~\cite{csqa} and SocialIQA~\cite{socialiqa}.
Although large-scale pre-trained models (PTMs)~\cite{bert,roberta} have surpassed human performance in a number of NLP benchmarks, it is still hard for PTMs to accurately capture and understand commonsense knowledge for accomplishing complex reasoning tasks~\cite{csqav2}.

In order to enhance the reasoning capacity, commonsense knowledge graphs~(CSKGs)~(\eg ConceptNet~\citep{conceptnet} and ATOMIC~\citep{atomic}) have been adopted for injecting external commonsense knowledge into PTMs.
By conducting entity linking to CSKGs, existing methods~\cite{Yasunaga2021qagnn,feng2020mhgrn} aim to capture the structured knowledge semantics via knowledge graph~(KG) encoders (\eg graph neural network~(GNN)~\cite{gat,gcn}), and then integrate the KG encoders for improving the commonsense reasoning capacity of PTMs~\cite{Yasunaga2021qagnn}. 

Despite the effectiveness, these approaches are built on highly complicated network architectures (involving both PTMs and GNNs).
Thus, it is difficult to explain how and why external commonsense knowledge improves the commonsense reasoning capacity of PTMs. 
Besides, existing CSKGs~\cite{mehrabi2021lawyers,nguyen2021refined} are mostly crowdsourced from massive selected resources~(\eg books, encyclopedias, and scraped web corpus), containing a wide variety of content. 
Without a clear understanding of how these external resources should be utilized, it is likely to incorporate irrelevant concepts or even knowledge biases~\cite{mehrabi2021lawyers,nguyen2021refined} into PTMs, which might hurt the reasoning performance. 
Indeed, some researchers have noted this issue and questioned whether existing GNN-based modules are over-complicated for commonsense reasoning~\cite{wang2021gsc}. 
Furthermore, they find that even a simple graph neural counter can outperform existing GNN modules on CommonsenseQA and OpenBookQA benchmarks.

However, existing studies can't well answer the fundamental questions about knowledge utilization for commonsense reasoning: How do external knowledge resources enhance the commonsense reasoning capacity of PTMs? What is necessarily required from external knowledge resources for PTMs?
Since the simplified knowledge-aware GNN has already yielded performance improvement on the CommonsenseQA~\cite{wang2021gsc}, we speculate that there might be a simpler solution if we could identify the essential knowledge for commonsense reasoning.

Focusing on this issue, we think about designing the solution by further simplifying the KG encoder. 
Based on our empirical analysis, we observe a surprising result that it is indeed \emph{relation features} from CSKGs, but not \emph{node features}, that are the key to the task of commonsense reasoning (See more details in Section~\ref{pilot}). 
According to this finding, we propose a rather simple approach to leveraging external knowledge resources for enhancing the commonsense reasoning capacity of PTMs. 
Instead of using a heavy GNN architecture, we design a lightweight KG encoder fully based on the multi-layer perceptron~(MLP), which utilizes \textbf{S}tatistical relation p\textbf{A}th from CSKGs as \textbf{FE}atures, namely \textbf{SAFE}.
We find that semantic relation paths can provide useful knowledge evidences for PTMs, which is the key information for helping commonsense reasoning. 
By conducting extensive experiments on five benchmark datasets, our approach achieves superior or competitive performance compared with state-of-the-art methods, especially when training data is limited. 
Besides the performance improvement, our approach largely reduces the parameters for encoding CSKGs (fewer than 1\% trainable parameters compared to GNN-based KG encoders~\citep{Yasunaga2021qagnn}). 

Our main contributions can be summarized as follows: (1) We empirically find that relation features from CSKGs are the key to the task of commonsense reasoning; (2) We design a simple MLP-based architecture with relation paths as features for enhancing the commonsense reasoning capacity of PTMs; (3) Extensive experiments conducted on five benchmark datasets demonstrate the effectiveness of our proposed approach, which also largely reduces the parameters of the KG encoder.

\section{Task Description}
According to pioneer works~\citep{csqa,obqa}, the commonsense reasoning task can be generally described as a multi-choice question answering problem: given a natural language question $q$ and a set of $n$ choices $\{c_{1},\cdots,c_{n}\}$ as the answer candidates, the goal is to select the most proper choice $c^{\star}$ from these candidates to answer the question based on necessary commonsense knowledge.

To explicitly capture commonsense knowledge, external commonsense knowledge graphs~(CSKGs) have often been utilized in this task, \eg ConceptNet~\citep{conceptnet}.
A CSKG can be formally described as a multi-relational graph $\mathcal{G} = (\mathcal{V}, \mathcal{R},\mathcal{E})$, where $\mathcal{V}$ is the set of all concept  (or entity) nodes (\eg \emph{hair} and \emph{water}), $\mathcal{R}$ is the set of relation types (\eg \emph{relatedto} and \emph{atlocation}), and $\mathcal{E} \subseteq \mathcal{V} \times \mathcal{R} \times \mathcal{V}$ is the set of relational links that connect two concept nodes in $\mathcal{V}$.

Following prior studies~\cite{lin2019kagnet}, we solve the commonsense reasoning task in a \emph{knowledge-aware} setting, where a CSKG $\mathcal{G}$ is available as input.  
We first link the mentioned concepts from the question and the answer candidates to the CSKG, so that we can leverage the rich semantic knowledge from the CSKG for commonsense reasoning. 
Based on the linked concepts in the question and each answer candidate, we further extract their neighbouring nodes from $\mathcal{G}$ and the relational links that connect them, to compose a subgraph $\mathcal{G}^{q,c_{i}}$ for characterizing  the commonsense knowledge about the question $q$ and the answer candidate $c_i$.

\ignore{The associated concepts in the question and candidates are denoted by  
$\mathcal{V}_q \subseteq \mathcal{V}$ and $\mathcal{V}_{c_{i}} \subseteq \mathcal{V}$, respectively.
Based on these seed nodes, we further extract the neighbouring nodes from $\mathcal{G}$ and the associated links that connect them, to compose a subgraph $\mathcal{G}_{(q,c_{i})}$ that characterizes the commonsense knowledge about the question $q$ and the answer candidate $c_i$.
}

\section{Empirical Analysis on the Commonsense KG Encoder} \label{pilot}
In this section, we conduct an empirical study to investigate how the external KG encoder helps PTMs with commonsense reasoning.

\subsection{Analysis Setup}
To conduct the analysis experiments, we select  QA-GNN~\citep{Yasunaga2021qagnn}, a representative approach that integrates PTM with GNN for the commonsense QA task, as the studied model.
We adopt the CommonsenseQA~\citep{csqa} and OpenBookQA~\cite{obqa}, two of the most widely used commonsense reasoning benchmarks, for evaluation, with the same data split setting in \cite{lin2019kagnet}.

We perform two analysis experiments: one examines the effect of the commonsense KG encoder, and the other one examines the effect of different features in the commonsense KG encoder.
To be specific, the two experiments focus on two key questions about commonsense reasoning:
(1) what is the effect of the commonsense KG encoder on PTMs?
(2) what is the key information within the commonsense KG encoder?

\subsection{Results and Findings}
Next, we conduct the experiments and present our findings of commonsense reasoning. 

\paratitle{Effect of Commonsense KG Encoder.} 
Since existing studies have widely utilized a GNN module to encode the commonsense knowledge, we examine its contribution to the improvement of reasoning performance. 
We consider comparing three variants of QA-GNN: (A) \emph{PTM-Only} that directly removes the GNN module and degenerates into a pure PTM, (B) \emph{PTM-Pred} that trains the PTM and GNN simultaneously but only makes the prediction with the PTM module, and (C) \emph{GNN-Pred} that trains the PTM and GNN simultaneously but only makes the prediction with the GNN module.

\begin{figure}[t]
	\centering
	\subfigure[CommonsenQA]{\label{fig-csqa-control}
		\centering
		\includegraphics[width=0.22\textwidth]{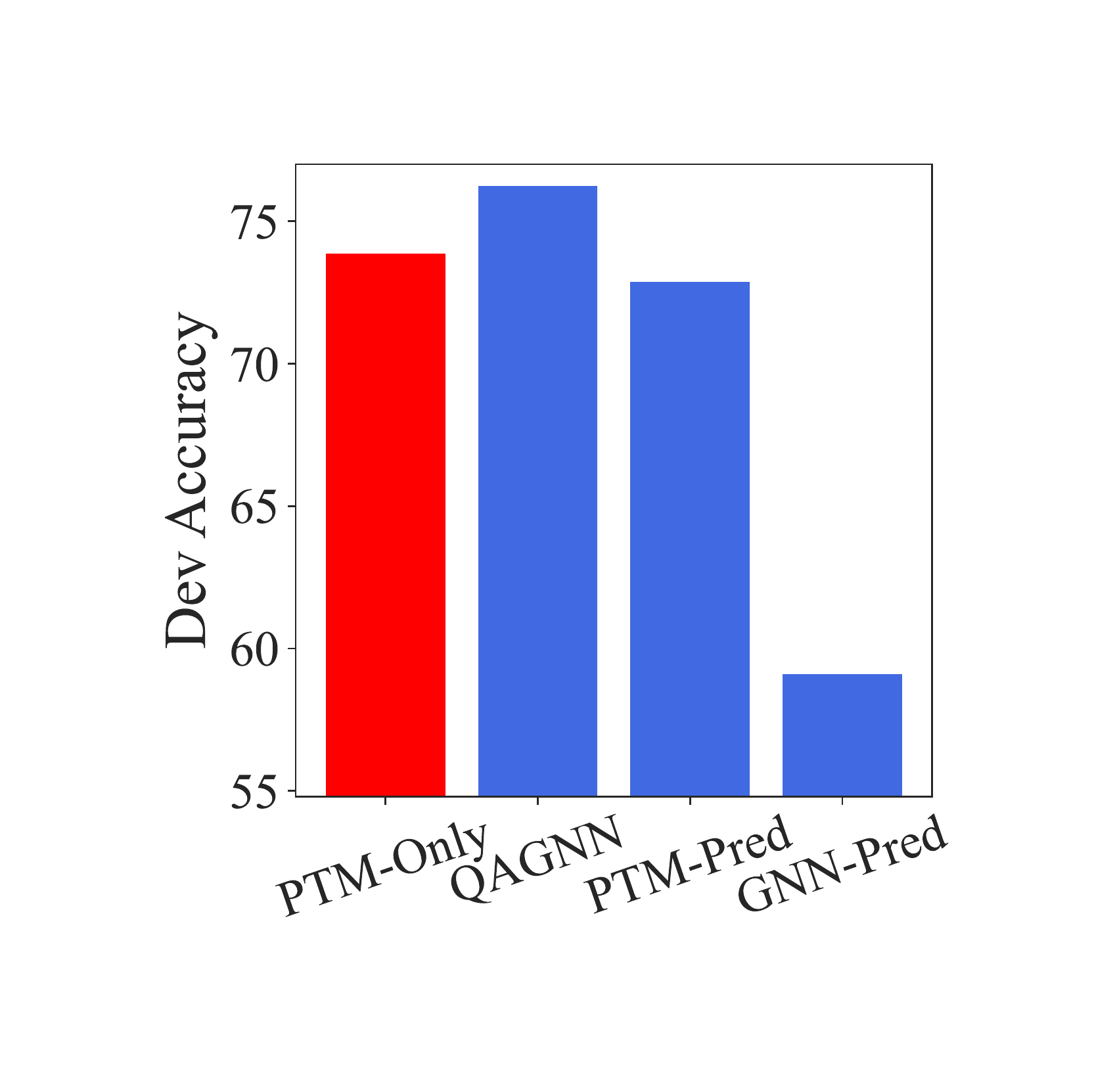}
	}
	\subfigure[OpenBookQA]{\label{fig-obqa-control}
		\centering
		\includegraphics[width=0.22\textwidth]{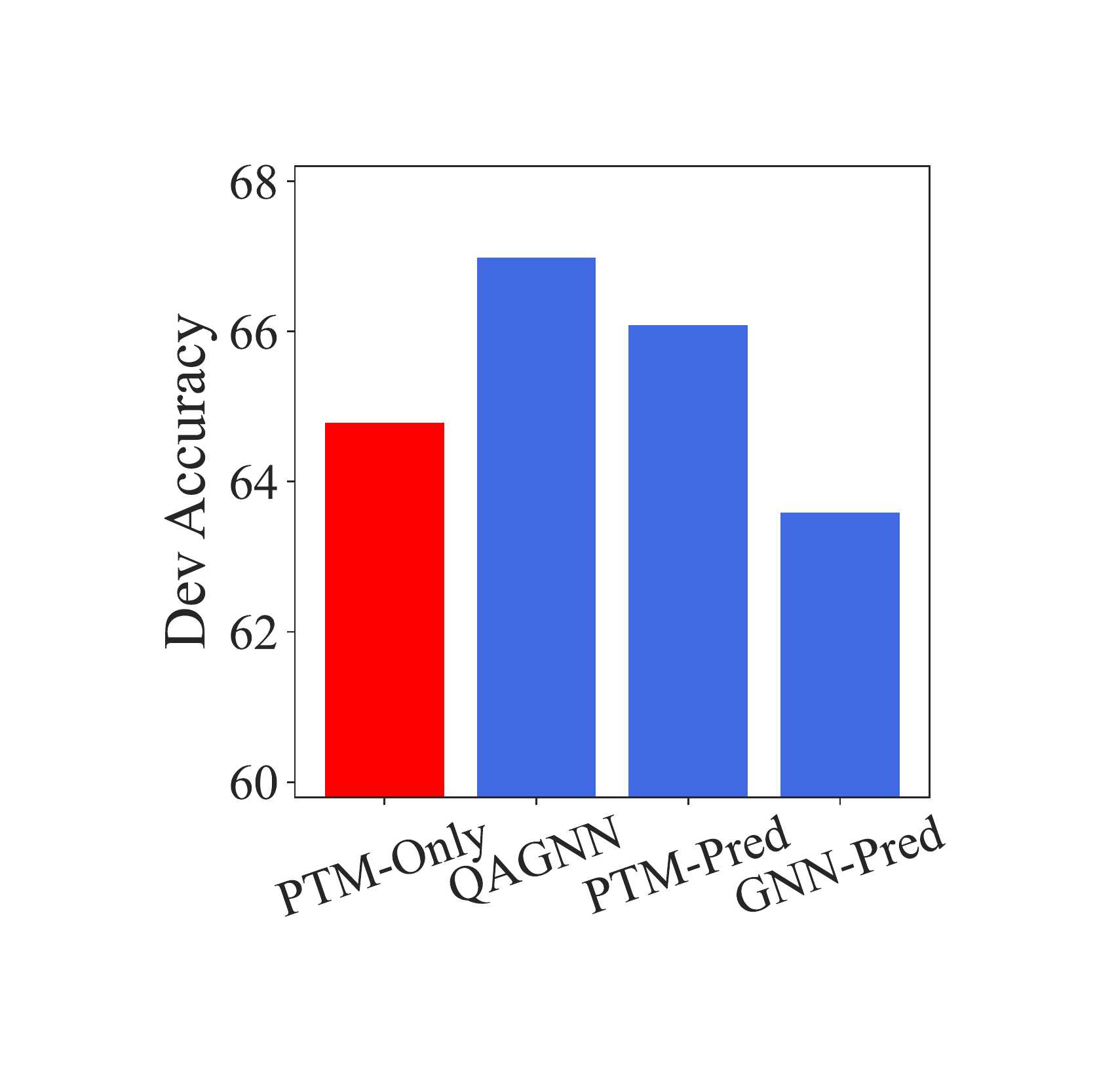}
	}
	\centering
	\caption{Performance comparison on CommonsenseQA and OpenBookQA (Dev accuracy).
	}
	\label{fig-kg-encoder}
\end{figure}

The comparison results are shown in Figure~\ref{fig-kg-encoder}.
As we can see,  using the predictions solely based on the GNN module~(\ie GNN-Pred) can only answer a relatively minor proportion of the questions (no more than 60\% in CommonsenseQA). 
As a comparison, when trained independently~(\ie PTM-Only) or jointly with the GNN module~(\ie PTM-Pred), the PTM module can answer a large proportion of the questions (at least 70\% in CommonsenseQA). 
Furthermore, the incorporation of the GNN encoder is useful to improve the performance of PTMs (PTM-Only \emph{v.s.} QAGNN). 
These results show that: 

$\bullet$~In the joint PTM-GNN approach, PTM contributes the most to the commonsense reasoning task, which is the key to the reasoning performance. 

$\bullet$~Commonsense KG encoder is incapable of performing effective reasoning independently, but can enhance PTM as the auxiliary role. 

\begin{figure}[t]
	\centering
	\subfigure{\label{fig-node-dimension}
		\centering
		\includegraphics[width=0.223\textwidth]{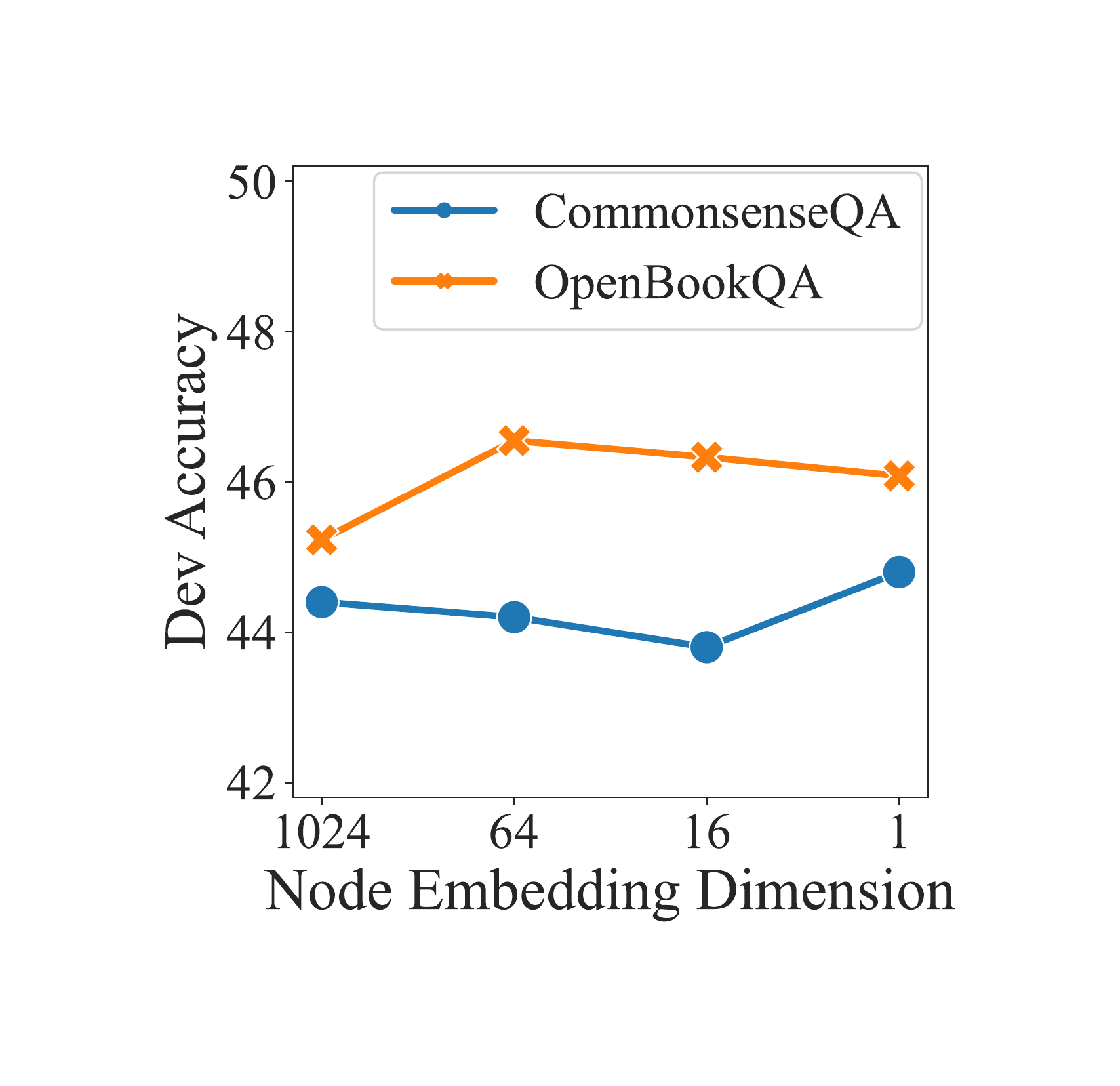}
	}
	\subfigure{\label{fig-edge-drop}
		\centering
		\includegraphics[width=0.215\textwidth]{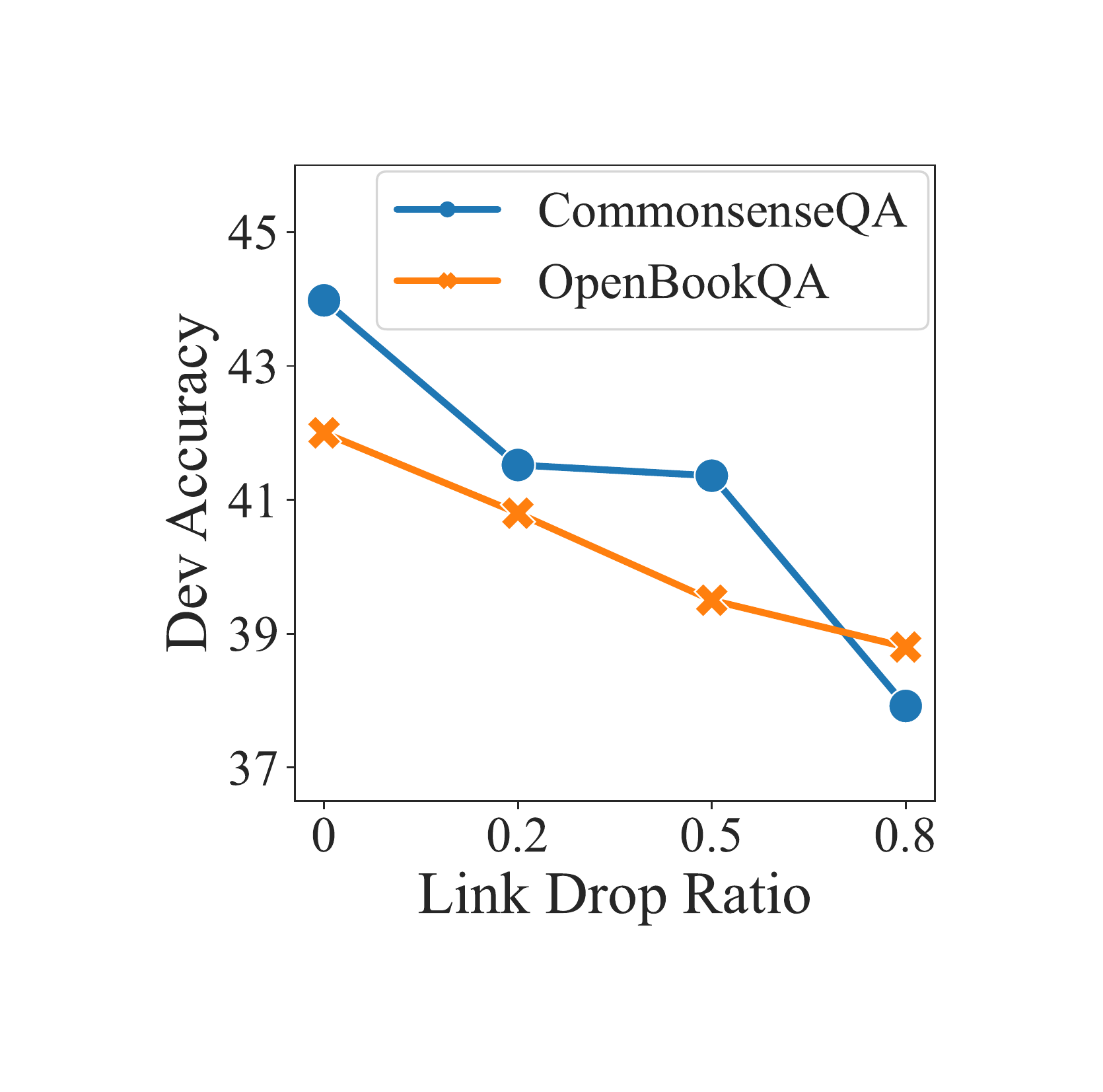}
	}
	\centering
	\caption{Performance examination for KG encoder on CommonsenseQA and OpenBookQA (Dev accuracy).
	}
	\label{fig-edge-node}
\end{figure}

\paratitle{Effect of Node/Relation Features from KG.} 
The major aim of the KG encoder is to characterize the commonsense knowledge and provide necessary knowledge evidence for enhancing the reasoning capacity of PTMs.
Generally, a CSKG consists of concept nodes and relational links. 
To identify the key knowledge information that is necessarily needed, we now examine the effect of node and relation features from CSKG. 
To eliminate the effect of PTM module, we remove it and compare the performance of only KG encoder under two experiment settings: (A) reducing the dimension of node embeddings to $d$ (PCA~\citep{pca} is applied to select $d$ most informative dimensions), and (B) randomly removing $p$ percent of relational links in the KG subgraph for a question-candidate pair.

As shown in Figure~\ref{fig-edge-node}, we surprisingly find that even after reducing the dimension of node embeddings to 1, the performance of the GNN encoder can be still improved. 
These results show that node features are not the key information utilized by the GNN encoder. 
In contrast, removing a considerable proportion of links significantly reduces the performance.
From these observations, we can conclude that: The relation features from the CSKG are indeed the key knowledge information that is actually needed by the KG encoder. 

\section{Approach}
\label{approach}
\begin{figure}[t]
  \centering
  \includegraphics[width=1\columnwidth]{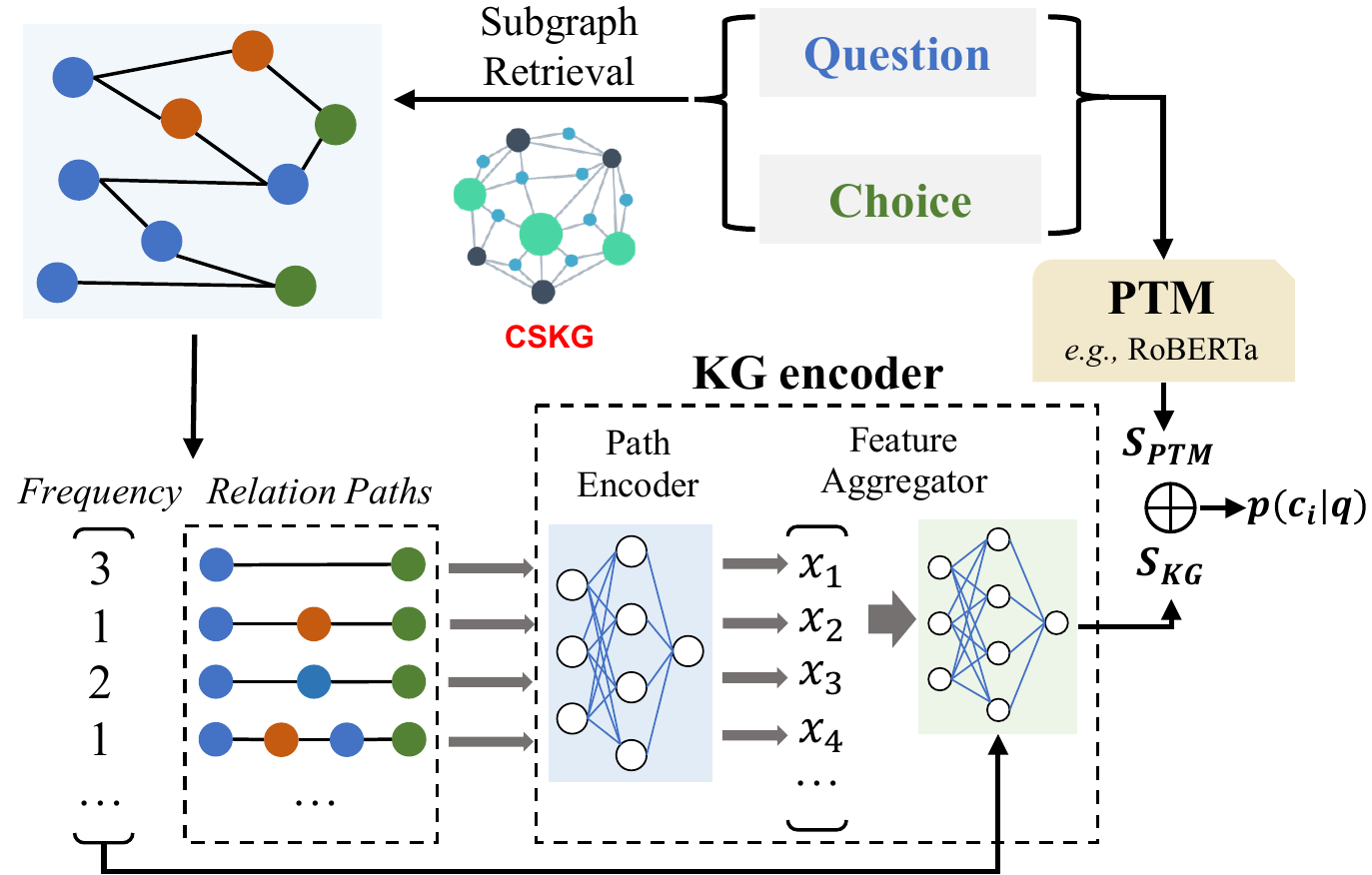}
  \caption{The illustration of our approach. We adopt an all-MLP KG encoder to model the extracted relation features from CSKG to enhance the PTM.
  }
  \label{fig:model}
\end{figure}

The former sections show that the role of the KG encoder on CSKGs is to mainly complement PTMs in the task of commonsense reasoning.
Instead of node features, relations features are the key to the KG encoder for improving PTMs.
Based on these findings, we develop a simple commonsense KG encoder based on the statistical relation features from CSKGs, namely \textbf{SAFE}.
Figure~\ref{fig:model} presents the overview of our model.

\subsection{Capturing High-Order Relation Semantics}
\label{extracting}
Since relation features are shown useful to improve the performance of commonsense reasoning, we consider extracting relation features for better capturing the knowledge semantics from the CSKG. 
Inspired by KG reasoning studies~\cite{lin2018multi,feng2020scalable}, we construct multi-hop relation paths that connect question nodes with answer candidate nodes on the CSKG to capture the higher-order semantic relatedness among them. 

Formally, given the commonsense subgraph $\mathcal{G}^{q,c_{i}}$ for the question $q$ and the answer candidate $c_{i}$, we first extract a set of relation paths within $k$ hops that connect a question concept node $v_q \in \mathcal{V}_q$ and an answer concept node $v_{c_i} \in \mathcal{V}_{c_i}$, denoted as $\mathcal{P}^{q,c_{i}}$.
Specifically, a path $p \in \mathcal{P}^{q,c_{i}}$ can be represented as a sequence of nodes and relations as $p=\{v_1, r_1, \cdots, r_{k-1}, v_k\}$.
Based on the empirical findings in Section~\ref{pilot}, we consider a simplified  representation for relation paths that removes node IDs but only keeps the relations on a path. 
To keep the role of each node, we replace a node ID by a three-valued type, indicating this node belongs to a \emph{question node} (0), \emph{answer node} (1) or \emph{others} (2).  
In this way, a path $p$ can be  represented by $p=\{ t_{v_1}, r_1, t_{v_2}, r_2, \cdots, r_{k-1}, t_{v_k}\}$, where $t_{v}$ is the role type of node $v$.
Since we remove explicit node IDs, our model can concentrate on more essential relation features.

Based on the above method, for a question $q$ and an answer candidate $c_i$, we extract all the simplified relation paths and count their frequencies among all the paths. 
We use $\mathcal{F}^{q,c_i}=\{ \langle p_j, f_j \rangle \}$ to denote all the paths for the question $q$ and the answer candidate $c_i$, where each entry consists of the $j$-th path $p_j$ and its frequency $f_j$.
Unlike prior approaches (\eg QA-GNN), we use such very simple features of relation paths from CSKGs to improve the reasoning capacity of PTMs.

\subsection{A MLP-based KG encoder}
\label{KG encoder}

Our KG encoder is built on a full MLP architecture based on simplified relation path features, consisting of a path encoder and a feature aggregator.

\paratitle{Path Encoder.} 
The path encoder is a two-layer MLP that encodes a relation path into a scalar feature value.
As shown in Section~\ref{extracting}, we can obtain the path feature set $\mathcal{F}^{q,c_i}=\{ \langle p_j, f_j \rangle \}$ for the question $q$ and the answer candidate $c_i$.
Different from general KGs, CSKGs usually contain much fewer types of relations (\eg 36 relations in ConceptNet), we adopt one-hot representations of these types to represent these relations. 
For node type (from \emph{question}, \emph{candidate} or \emph{others}), we also adopt the similar representations. 
Then, we concatenate these one-hot vectors to compose the sparse representation of a relation path $p$ in order, denoted as $\mathbf{v}_{p}$. 
Subsequently, the sparse path representation is encoded by a two-layer MLP~(\ie the path encoder) to produce the corresponding scalar feature value $x_{p}$:
\begin{equation}
    x_{p}=\text{MLP}_{2}(\text{MLP}_{1}(\mathbf{v}_{p})),
\end{equation}
where $x_{p}$ reflects the importance of such a relation path for commonsense reasoning.

\paratitle{Feature Aggregator.}
Based on the above path encoder, we can generate the scalar feature values for all the relation paths in the feature set $\mathcal{F}^{q,c_i}=\{ \langle p_j, f_j \rangle \}$.
The feature aggregator aims to aggregate these feature values to produce the confidence score of  the answer candidate \emph{w.r.t.} the question, from the KG perspective.
Concretely, we sum the different feature values of relation paths weighted by their frequencies as follows: 
\begin{equation}
    x_{q,c_i}=\sum_{\langle p_j, f_j\rangle \in \mathcal{F}^{q,c_i}}x_{p_j} \cdot f_j,
\end{equation}
where $x_{p_j}$ is the mapping feature value of path $p_j$ and $f_j$ is the frequency of path $p_j$. Here, $x_{q,c_i}$ aims to capture the overall confidence score based on the subgraph $\mathcal{G}^{q,c_i}$ given the question and the answer candidate.
However, since the weighted sum is likely to cause extreme values (\ie too large or too small), we  add an extra two-layer MLP for scaling:
\begin{equation}
    S_{KG}(q,c_i)=\text{MLP}_{4}(\text{MLP}_{3}(x_{q,c_i})),
\label{skg}
\end{equation}
where $S_{KG}$ is the prediction score indicating the confidence level that candidate $c_i$ is the right answer to question $q$ from the perspective of KG. 

\subsection{Integrating KG Encoder with PTM}

In this part, we integrate the above KG encoder with the PTM for commonsense reasoning. 

\paratitle{The PTM Encoder}. Following existing works~\cite{Yasunaga2021qagnn}, we utilize a PTM  as the backbone of commonsense reasoning. 
Given a question $q$ and an answer candidate $c_i$, we concatenate their text to compose the input of the PTM. 
After encoding by the multiple Transformer layers, we select the output of the \texttt{[CLS]} token in the last layer as the contextual representation of the question-candidate pair, denoted by $\mathbf{h}_{cls}$.
Then, we feed $\mathbf{h}_{cls}$ into a MLP layer to produce a scalar output $S_{PTM}$,
\begin{align}
\mathbf{h}_{cls}&=\text{PTM}(q,c_i),  \\
S_{PTM}(q,c_i) &= \text{MLP}(\mathbf{h}_{cls}),
\label{sptm}
\end{align}
which is the plausibility score of the answer candidate from the perspective of PTM.

\paratitle{Combining the Prediction Scores}. 
We then derive the prediction score of each answer candidate for a question by leveraging both the PTM and KG encoder based on either textual or structured semantics.
For each question-candidate pair $(q, c_i)$, we combine the prediction scores of the two modules as: 
\begin{equation}
   S(q,c_i) = S_{PTM}(q,c_i) + S_{KG}(q,c_i),
\end{equation}
where $S_{PTM}(q,c_i)$ (Eq.~\ref{sptm}) and $S_{KG}(q,c_i)$ (Eq.~\ref{skg}) are the prediction scores of PTM and KG encoder, respectively. 
Given a set of answer candidates $\{c_1,...,c_n\}$, we further normalize $S(q,c_i)$ into a conditional probability $\text{Pr}(c_i|q)$ via the softmax operation over the $n$ candidates. 

During the training stage, we optimize the parameters of the whole model~(including both the PTM and KG encoder) with the cross entropy loss between the predictions and the ground-truth answer (based on the probability distribution $\{ \text{Pr}(c_i|q)\}_{i=1}^n$  ).
During inference, we first compute the probability score $\text{Pr}(c_i|q)$ for each answer candidate, and then select the highest one as the predicted answer.

\begin{table}[t]
    \centering
    \small
     \begin{tabular}{lccc|c}
      \toprule
            & \textbf{RGCN} & \textbf{MHGRN}  &  \textbf{QAGNN} & \textbf{SAFE}  \\
      \midrule
       Node emb. & $\surd$  & $\surd$ & $\surd$ &$\times$ \\
       Relation & $\surd$ & $\surd$ & $\surd$ & $\surd$\\
       GNN & $\surd$ & $\surd$ & $\surd$ & $\times$ \\
       MLP-based &$\times$  &$\times$  & $\times$ &$\surd$\\
      \midrule
      \# Params & 365K & 547K & 2845K & 4.7k\\
    \bottomrule
    \end{tabular}
    \caption{Comparisons of different KG encoders for commonsense reasoning. Instead of using node embeddings and GNN structure, we adopt relation paths as the input features and incorporate a full MLP architecture.
    } 
    \label{tab:comparsion}
\end{table}

\subsection{Comparison with Existing KG Encoders}
For the task of commonsense reasoning, it has become a common approach  by integrating PTM with an external KG encoder based on CSKGs. 
The major difference among these methods (including our approach) lies in the design of the KG encoder. 
Next, we compare these variants for the KG encoder. 

We summarize the comparison between our KG encoder and representative KG encoders in Table~\ref{tab:comparsion}.
We can see that, our approach no longer lies in the node embeddings and the structure of GNNs. 
Instead, we mainly utilize relation paths as the features of the KG encoder, which is built on a simple MLP-based architecture. 
Therefore, the number of the model parameters involved in our KG encoder is much smaller than those of existing KG encoders. 
As will be shown in Section~\ref{experiment}, our KG encoder yields better or at least comparable performance compared with existing GNN-based encoders, based on the same configuration for PTMs. 

Specifically, our approach can largely reduce the computational costs for encoding the CSKG. 
For our approach, we need to extract the relation paths from question nodes to all the answer candidate nodes on the CSKG, and it can be efficiently fulfilled via a $k$-hop constrained Depth-First Search~\cite{dfs}, which can be pre-computed in offline processing.  
When the relation paths have been extracted, it is efficient to encode these paths with our MLP architecture. 
Such a process can be easily paralleled or accelerated by optimized matrix multiplication.
In contrast, existing GNN-based encoders rely on iterative propagation and aggregation on the entire subgraph, which takes a much larger computational time cost.

\section{Experiment} \label{experiment}

\subsection{Experimental Setup}
In this part, we introduce the experimental setup.

\paratitle{Evaluation Tasks.}
We conduct experiments on five commonsense reasoning tasks, shown in Table~\ref{tab:number_static}.

$\bullet$ \textbf{CommonsenseQA}~\citep{csqa} is a 5-way multiple-choice QA dataset. It is created based on ConceptNet~\citep{conceptnet}.

$\bullet$ \textbf{OpenBookQA}~\citep{obqa} is a 4-way multiple-choice QA dataset about elementary science questions to evaluate the science commonsense knowledge.

$\bullet$ \textbf{SocialIQA}~\citep{socialiqa} is a 3-way multiple-choice QA dataset to evaluate the understanding of social commonsense knowledge.

$\bullet$ \textbf{PIQA}~\citep{piqa} is a binary-choice QA dataset about physical commonsense.

$\bullet$ \textbf{CoPA}~\citep{copa} is a commonsense inference dataset, to select the most plausible alternative with the causal relation to the premise.

\begin{table}[t]
\centering
\small
\begin{tabular}{l|rrr}
\toprule
\textbf{Task} & \textbf{Train} & \textbf{Dev} & \textbf{Test} \\ 
\midrule
CommonsenQA & 9,741 & 1,221 & 1,140 \\
OpenBookQA & 4,957 & 500 & 500 \\
SocialIQA & 33,410 & 1,954 & - \\
PIQA & 16,113 & 1,838 & - \\
CoPA & - & 500 & 500 \\
\bottomrule
\end{tabular}
\caption{Statistics of the datasets. ``-'' denotes the unused or not available dataset split in our experiments.}
\label{tab:number_static}
\end{table}

\paratitle{Data Preprocessing.}
For CommonsenseQA and OpenBookQA, we use their original train/dev/test split settings.
Since the test set of CommonsenseQA is not available, we follow previous work~\cite{lin2019kagnet} that extracts 1,241 examples from the original training set as the test set.
Besides, the test sets of SocialIQA and PIQA are not available. 
Therefore, we report the experimental results on their development sets for a fair comparison~\cite{shwartz2020unsupervised}.
For CoPA that only provides development and test sets, we follow \citet{niu2021sematic} to train models on the development set and evaluate the performance on the test set.
For commonsense KG, we adopt \emph{ConceptNet}~\cite{conceptnet}, a general-domain and task-agnostic CSKG, as our external knowledge source $\mathcal{G}$ for all the above models and tasks.
For each question-candidate pair $(q, c_{i})$, we follow previous works~\cite{lin2019kagnet,feng2020mhgrn} to retrieve and construct the subgraph $\mathcal{G}^{q,c_{i}}$ from the CSKG $\mathcal{G}$.

\paratitle{Baseline Methods.}
We compare our model with the following six baseline methods, including a fine-tuned PTM and five PTM+GNN models:

$\bullet$ \textbf{Fine-tuned PTM} directly fine-tunes a PTM without using any CSKG. 
We use RoBERTa-large~\cite{roberta} for all tasks.
Additionally, we also use BERT-large~\citep{bert} and AristoRoBERTa~\citep{aroberta} for OpenBookQA to evaluate the generality of our KG-encoder. 

$\bullet$ \textbf{PTM+GNN models} integrate PTM with additional GNN-based KG encoders. Based on the same PTM (the above baseline), we consider five variants with different KG encoders: 
(1) \emph{Relation Network}~(RN)~\citep{santoro2017simple} using a relational reasoning structure over the CSKG; 
(2) \emph{GcoAttn}~\citep{lin2019kagnet} using a graph concept attention model to aggregate entity information from the CSKG;
(3) \emph{RGCN}~\citep{rgcn} extending the GCN with relation-specific weights;
(4) \emph{MHGRN}~\citep{feng2020mhgrn} using a GNN architecture reasoning over the CSKG that unifies both GNNs and path-based models;
(5) \emph{QA-GNN}~\citep{Yasunaga2021qagnn} using a GAT to perform jointly reasoning over the CSKG.

For all these methods, we adopt the same architecture and configuration for the PTM, so that we can examine the effect of different KG encoders. 

\begin{table*}[t]
\centering
\small
\begin{tabular}{p{0.262\columnwidth}cccccccccccc}
    \toprule
        \multirow{2}{*}{\textbf{Methods}}&
        \multicolumn{6}{c}{CommonsenseQA} & \multicolumn{6}{c}{OpenBookQA}\\
        \cmidrule(lr){2-7} \cmidrule(lr){8-13}
        & 5\% & 10\% & 20\% & 50\% & 80\% & 100\% & 5\% & 10\% & 20\% & 50\% & 80\%  & 100\%\\
        \midrule
        RoBERTa-large & 29.66 & 42.84 & 58.47 & 66.13 & 68.47 & 68.69$^{\dag}$ & 37.00 & 39.4 & 41.47 & 53.07 & 57.93 & 64.8$^{\dag}$\\
        \midrule
        + RGCN & 24.41 & 43.75 & 59.44 & 66.07 & 68.33 & 68.41$^{\dag}$ & 38.67 & 37.53 & 43.67 & 56.33 & 63.73 & 62.45$^{\dag}$\\
        + GconAttn & 21.92 & 49.83 & 60.09 & 66.93 & 69.14 & 68.59$^{\dag}$ & 38.60 & 36.13 & 43.93 & 50.87 & 57.87 & 64.75$^{\dag}$\\
        + RN & 23.77 & 34.09 & 59.90 & 65.62 & 67.37 & 69.08$^{\dag}$ & 33.73 & 35.93 & 41.40 & 49.47 & 59.00 & 65.20$^{\dag}$\\
        + MHGRN & 29.01 & 32.02 & 50.23 & 68.09 & 70.83 & 71.11$^{\dag}$ & 38.00 & 36.47 & 39.73 & 55.73 & 55.00 & 66.85$^{\dag}$\\
        + QA-GNN & 32.95 & 37.77 & 50.15 & 69.33 & 70.99 & 73.41$^{\dag}$ & 33.53 & 35.07 & 42.40 & 54.53 & 52.47 & 67.80$^{\star}$\\
        \midrule
        + SAFE(\textbf{Ours}) & \textbf{36.45} & \textbf{56.51} & \textbf{65.16} & \textbf{70.72} & \textbf{73.22} & \textbf{74.03} & \textbf{38.80} & \textbf{41.20} & \textbf{44.93} & \textbf{58.33} & \textbf{65.60} & \textbf{69.20}\\
    \bottomrule
\end{tabular}
\caption{Performance comparison on CommonsenseQA and OpenBookQA with different proportions of training data. We report the average test performance of three runs, and the best results are highlighted in bold. $\dag$ indicates the reported results from \citet{Yasunaga2021qagnn}. $\star$ indicates the reported results from \citet{wang2021gsc}}
\label{tab:few-shot}
\end{table*}

\begin{table}[t]
\centering
\small
\begin{tabular}{lcccc}
    \toprule  
    \textbf{Methods}& \textbf{SocialIQA} & \textbf{PIQA} & \textbf{CoPA}\\
    \midrule  
    RoBERTa-large  & 78.25 & 77.53 & 67.60\\
    \midrule
    + {GcoAttn} & 78.86 & 78.24 &70.00 \\
    + {RN} & 78.45 & 76.88 &70.20 \\
    + {MHGRN} & 78.11 & 77.15 & 71.60\\
    + {QAGNN} & 78.10 & 78.24 & 68.40 \\
    \midrule
    + {SAFE}~(\textbf{Ours}) &  \textbf{78.86}  & \textbf{79.43} &  \textbf{71.60} \\
    \bottomrule 
\end{tabular}
\caption{Performance comparison on SocialIQA, PIQA, and CoPA (Dev accuracy).
}
\label{tab:other_qa}
\end{table}

\subsection{Implementation Details}
We implement all PTMs based on HuggingFace Transformers~\cite{huggingface}.
For all the baselines, we keep the common hyper-parameters as identical as possible and set their special hyper-parameters following the suggestions from the original papers.
In our approach, we extract the relation paths with no more than 2 hops between the concept nodes from the question and the answer candidate.
We tune the hidden dimension of MLPs from the path encoder in \{32, 64, 100\}, and the batch size in \{32, 48, 60, 120\}. 
The parameters of the model are optimized by RAdam~\citep{radam}, and the learning rate of the PTM and the KG encoder is also tuned in \{1$e$-4, 1$e$-5, 2$e$-5\} and \{1$e$-3, 1$e$-2\}, respectively.
To accelerate the training process, we don't incorporate Dropout regularization in our model.
All the above hyper-parameters are tuned on the development set.

\subsection{Results Analysis}
Following previous works~\cite{Yasunaga2021qagnn,wang2021gsc}, we take the results on CommonsenseQA and OpenBookQA as the main experiments to compare different methods. 
In order to test their robustness to data sparsity, we examine the performance under five different proportions of training data, \ie $\{5\%, 10\%, 20\%, 50\%, 80\%, 100\%\}$.

\paratitle{CommonsenseQA and OpenBookQA.}
The results of different methods on CommonsenseQA and OpenBookQA are presented in Table~\ref{tab:few-shot}.

Comparing the results under the full-data setting~(\ie 100\% training data), we can see that all the PTM+GNN methods perform better than vanilla PTM (\ie RoBERTa-large).
It indicates that the KG encoder on the CSKG is able to incorporate useful knowledge information to improve PTMs on commonsense reasoning tasks.
Additionally, among all the PTM+GNN baselines, QA-GNN performs the best. The major reason is that QA-GNN uses the PTM to estimate the importance of KG nodes and connects the QA context and the CSKG to form a joint graph, which is helpful to improve the reasoning ability on the CSKG.
Finally, our method consistently outperforms all the baselines. Our approach incorporates a lightweight MLP architecture as the KG encoder with relation paths as features.
It reduces the parameter redundancy of the KG encoder and focuses on the most essential features for reasoning, \ie semantic relation paths. 
Such an approach is effective to enhance the commonsense reasoning capacity of PTMs.

Comparing the results under different sparsity ratios of training data, we can see that the performance substantially drops when the size of training data is reduced.
While, our method performs consistently better than all baselines. 
It is because that our KG encoder consists of significantly fewer parameters than those of the baselines, which reduces the risk of overfitting and endows our approach with better robustness in data scarcity scenarios.

\begin{table}[tb]
\centering
\small
\begin{tabular}{lcc}
\toprule
\textbf{Methods}          & \textbf{BERT-large}     & \textbf{AristoRoBERTa}     \\
\midrule
Fine-tuned PTMs    & 59.00  & 78.40$^{\dag}$      \\
\midrule
+ RGCN  & 45.40    & 74.60$^{\dag}$ \\
+ GconAttn & 48.20    & 71.80$^{\dag}$ \\
+ RN & 48.60    & 75.35$^{\dag}$ \\
+ MHGRN & 46.20 & 80.60$^{\dag}$        \\
+ QA-GNN & 58.47   & 82.77$^{\dag}$ \\
\midrule
+ SAFE~(\textbf{Ours})   & \textbf{59.20}   &  \textbf{87.13}  \\
\bottomrule
\end{tabular}
\caption{Evaluation  with other PTMs on OpenBookQA (average test accuracy of three runs). Methods with AristoRoBERTa use the textual evidence by \citet{science-exam} as an additional input to the QA context. \dag ~indicates reported results in \cite{Yasunaga2021qagnn}.}
\label{tab:obqa_main}
\end{table}

\paratitle{Other Commonsense Reasoning Datasets}.
To further verify the effectiveness of our method, we also compare the results of different methods on other commonsense reasoning datasets.
These datasets are from different domains or different tasks.
These results are shown in Table~\ref{tab:other_qa}.
Similarly, our approach also achieves the best performance in most cases. 
It indicates that our approach is generally effective for various commonsense reasoning datasets or tasks, by outperforming competitive but complicated baselines.
Among all the datasets, our approach improves the performance of the PTM on CoPA dataset by a large margin. 
The reason is that CoPA is a small dataset with only 500 training examples. 
Baselines with heavy architectures are easy to overfit on it.
In contrast, our KG encoder is lightweight, which is more capable of resisting the overfitting issue.

\subsection{Evaluation with Other PTMs}
The major contribution of our approach lies in the lightweight KG encoder, which can be also used to enhance the commonsense reasoning capacity of various PTMs. 
To validate it, we examine the performance of our KG encoder when integrated with two other PTMs, \ie BERT-large and AristoRoBERTa, on OpenBookQA dataset.

As shown in Table~\ref{tab:obqa_main}, the BERT-large and AristoRoBERTa enhanced by our KG encoder perform better than original PTMs. 
Especially, our KG encoder can improve the performance of AristoRoBERTa by a large margin (with 8.73\% improvement). 
These results show that our KG encoder is a general method to improve PTMs for commonsense reasoning. 
In contrast, when adapting other KG encoders to these two PTMs, the performance decreases in most cases. 
It is mainly because these KG encoders have complicated architectures, which may not be easily adapted to other PTMs.  

\subsection{Hyper-parameters Analysis}
\begin{figure}[t]
\small
  \centering
  \includegraphics[width=0.62\columnwidth]{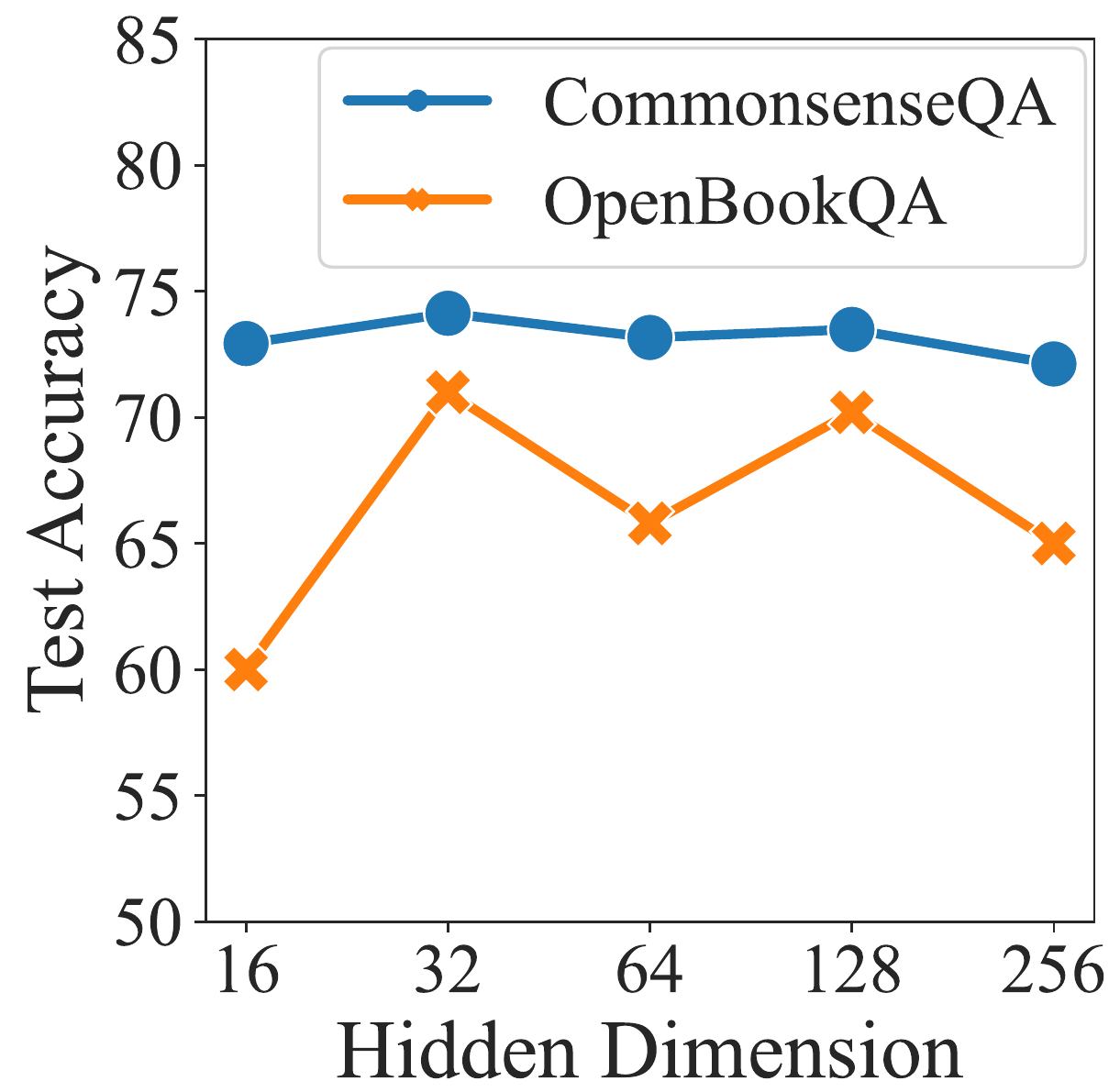}
  \caption{Analysis of different hidden dimension size of our SAFE model.}
  \label{fig-hidden}
\end{figure}

For hyper-parameter analysis, we study the hidden dimension size of the MLP in the path encoder.
Concretely, we evaluate our model with varying values of the hidden dimension size on CommonsenseQA and OpenBookQA datasets using RoBERTa-large model. 
The results are shown in Figure~\ref{fig-hidden}.
We can see that with the increase of the hidden dimension size, the performance improves at first and then drops to some extent. 
The possible reason lies in two aspects.
On the one hand, a too small hidden dimension size makes the path encoder hard to represent sufficient information from relation paths for commonsense reasoning.
On the other hand, a larger hidden dimension size enlarges the parameter number of our KG encoder, which increases the risk of overfitting that may cause performance degradation.

\subsection{Case Study}
\begin{figure}[t]
  \centering
  \includegraphics[width=0.85\columnwidth]{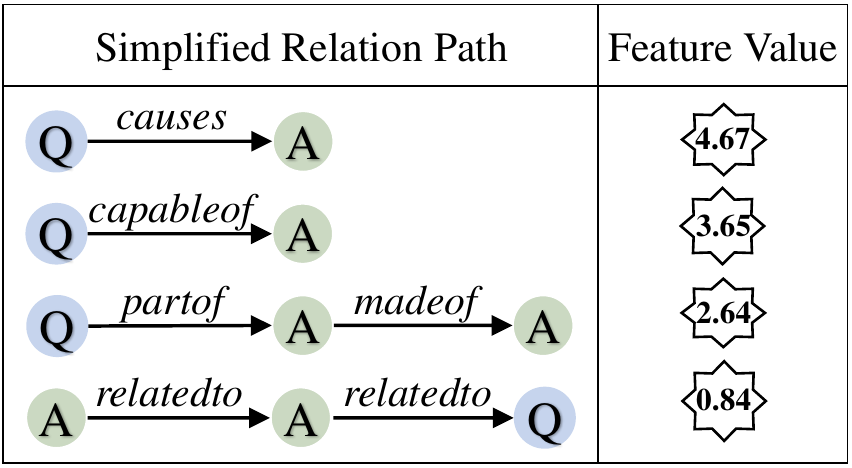}
  \caption{The generated feature values of relation path examples by the path encoder. \textbf{Q} and \textbf{A} denote the concept nodes from the question and the answer candidate, respectively.}
  \label{fig:case_study}
\end{figure}

We propose a rather simple KG encoder to effectively utilize the relation features from the CSKG, which first computes the feature values of the relation paths and then aggregates these values as the confidence score of the question and choice from the perspective of KG.
In this way, we can generate a table in advance that maps each type of relation path into its feature value that reflects its contribution to the confidence score.
Based on this table, it is convenient to directly judge the importance of the relation path and quickly assess the confidence about if the choice is the answer to the question from the perspective of KG.
Figure~\ref{fig:case_study} shows some path-value examples on CommonsenseQA dataset.
As we can see, the path with a higher value indeed provide more persuasive evidence (\eg \emph{causes} and \emph{capableof}) that indicates the choice is more likely to be the answer to the question.
In contrast, the path with a lower value usually represents an ambiguous relationship~(\eg \emph{relatedto}), which contributes less to the judge of whether the choice is the answer.

\section{Related Work} \label{related_work}
We review the related studies in two aspects, \ie commonsense reasoning and KG-enhanced pretrained models.

\paratitle{Commonsense Reasoning.}
Commonsense reasoning tasks aim to evaluate the understanding of commonsense knowledge~\cite{davis2015commonsense}, \eg physical commonsense~\citep{hswag}, which are mostly formulated as a multi-choice QA problem.
Early studies either rely on explicit text features~\citep{pmi} to capture the relations between the question and answer candidates, or adopt neural networks~(\eg DNN or LSTM)~\citep{dnn4qa,lstm4qa} to model the implicit correlation features.
Recently, pre-trained models (PTM)~\citep{bert, roberta} have achieved remarkable performance on commonsense reasoning tasks.
Furthermore, a surge of works incorporate external knowledge resources to further improve the reasoning performance.
Among them, CSKG~(\eg~ ConceptNet~\citep{conceptnet}) has been widely studied, and existing works mainly adopt graph neural networks to learn useful commonsense knowledge from the CSKG to enhance PTMs.
Based on these works, we systemically study what is necessarily needed from CSKGs for improving PTMs. Our analysis leads to an important finding that relation features mainly contribute to the performance improvement, and we design a lightweight MLP architecture to simplify the KG encoder. 

\paratitle{KG-Enhanced Pre-trained Models.}
Recently, a series of works focus on enhancing PTMs with external KGs to improve the performance on factual knowledge understanding~\cite{colake,wang2021kepler} and knowledge reasoning tasks~\cite{csqa,zhang2019ernie,he-kgc-www}.
These works inject the structured knowledge from the external KG into PTMs in either pre-training or fine-tuning stage. 
The first class of works mainly focus on devising knowledge-aware pre-training tasks~\cite{wang2021kepler,zhang2019ernie} to improve the understanding of entities or triples from the KG, \eg knowledge completion~\cite{wang2021kepler} and denoising entity auto-encoder~\cite{zhang2019ernie}.
Another class of works adopt task-specific KG encoders to enhance PTMs during fine-tuning, \eg path-based relation network~\cite{feng2020mhgrn} and GNN~\cite{Yasunaga2021qagnn}.
Different from them, we aim to directly enhance PTMs with a KG encoder on the downstream commonsense reasoning tasks, and design a rather simple yet effective KG encoder.

\section{Conclusion} \label{conclusion}
In this work, we study how the external commonsense knowledge graphs~(CSKGs) are utilized to improve the reasoning capacity of pre-trained models~(PTMs). 
Our work makes an important contribution to understanding and enhancing the commonsense reasoning capacity of PTMs. 
Our results show that relation paths from the CSKG are the key to performance improvement. 
Based on this finding, we design a rather simple MLP-based KG encoder with relation paths from the CSKG as features, which can be generally integrated with various PTMs for commonsense reasoning tasks.
Such a lightweight KG encoder has significantly fewer than 1\% trainable parameters compared to previous GNN-based KG encoders.
Experimental results on five commonsense reasoning datasets demonstrate the effectiveness of our approach.

In future work, we will study how to effectively leverage the commonsense knowledge from large-scale unstructured data to improve PTMs.
We will also try to apply our approach to other knowledge-intensive tasks, \eg knowledge graph completion and knowledge graph based question answering~\cite{lan-kbqa-ijcai}.

\section{Ethical Consideration} \label{ethical}
This work primarily investigates how external commonsense knowledge graphs (CSKGs) enhance the commonsense reasoning capacity of pre-trained models (PTMs) and proposes a simple but effective KG encoder on CSKGs to enhance PTMs.
A potential problem derives from using PTMs and CSKGs in our approach.
PTMs have been shown to capture certain biases from the data that have been pre-trained on~\cite{bender2021dangers}. 
And existing works~\cite{mehrabi2021lawyers} have found that CSKGs are likely to contain biased concepts derived from human annotations.
However, a comprehensive analysis of such biases is outside of the scope of this work.
It is a compelling direction to investigate to what extent the combination of CSKGs and PTMs can help mitigate such biases.
An alternative consideration is to consider filtering biased concepts in the process of subgraph extraction from the CSKG. 
By devising proper rules, it is promising to reduce the influence of biased concepts on our approach.

\section{Acknowledgments}
This work was partially supported by Beijing Natural Science Foundation under Grant No. 4222027, and  National Natural Science Foundation of China under Grant No. 61872369, Beijing Outstanding Young Scientist Program under Grant No. BJJWZYJH012019100020098.
This work is also supported by Beijing Academy of Artificial Intelligence~(BAAI). Xin Zhao is the corresponding author.

\bibliography{custom}
\bibliographystyle{acl_natbib}

\clearpage

\end{document}